\documentclass[runningheads]{llncs}

\usepackage[T1]{fontenc}
\usepackage{graphicx}
\usepackage{booktabs}
\usepackage{multirow}
\usepackage{amsmath}
\usepackage{xcolor}
\usepackage[hidelinks]{hyperref}
\usepackage{esvect}

\begin{document}

\title{Towards Grounded GI Endoscopy VQA via Multi-Task Learning on Small VLMs}
\titlerunning{Towards Grounded GI VQA on Small VLMs}

\author{Itbaan Safwan\inst{1}\thanks{Corresponding author} \and
Ramail Khan\inst{1} \and
Muhammad Annas Shaikh\inst{1} \and
Muhammad Atif Tahir\inst{1}}
\authorrunning{I. Safwan et al.}
\institute{Institute of Business and Administration, Karachi, Pakistan\\
\email{\{i.safwan.26197, r.khan.26924, m.shaikh.26919\}@khi.iba.edu.pk,
atiftahir@iba.edu.pk}}

\maketitle

\begin{abstract}
Gastrointestinal (GI) endoscopic image analysis has shifted from single-label classification toward visual question answering (VQA), where a model must answer free-form clinical questions about an image. While recent vision-language models (VLMs) achieve promising answer accuracy on this task, clinical adoption also requires the model's internal representations to reflect the visual evidence behind its answers. We propose a simple multi-task fine-tuning recipe that constructs auxiliary grounding and description tasks from an existing VQA dataset with minimal additional annotation: expert-annotated polyp masks are reused directly, while a GI-domain pretrained classifier with Grad-CAM localization provides weak supervision for finding categories that lack ground-truth masks. Three small VLM backbones are fine-tuned with low-rank adaptation under matched VQA-only and multi-task recipes on Kvasir-VQA-x1, and we show consistent accuracy gains together with improved implicit alignment between answer tokens and the relevant image region, evaluated on both in-distribution and out-of-distribution data.

\keywords{Vision Language Models \and Weak Supervision \and Multi-Task Learning \and Medical VQA \and Visual Grounding \and Parameter-Efficient Fine-Tuning}
\end{abstract}

\section{Introduction}

GI endoscopy generates large volumes of imagery that must be reviewed under time pressure, motivating automated tools to support clinical decision-making. Early automated approaches framed the problem as classification: detecting polyps, identifying landmarks, or flagging specific findings in a single frame \cite{pogorelov2017kvasir}. While useful for screening, this framing is limited to the narrow set of questions each classifier was built for.

More recently, VQA has emerged as a richer interface for GI image analysis \cite{Gautam2025Jun}. A single VQA model can answer diverse clinical questions about the same image, covering finding presence, type, color, size, and location. Vision-language models (VLMs) have been adapted to this setting with promising accuracy \cite{khanal2025hallucination}, though existing evaluations predominantly involve large backbones of 4B parameters or more \cite{Gautam2025Jun}, leaving it unclear whether similar benefits transfer to smaller, resource-constrained models. A model that produces correct answers by memorizing statistical associations rather than attending to the relevant image region is difficult to trust or audit. This concern is well documented in the broader VQA literature \cite{das2016human} and remains an active challenge in medical settings specifically, where recent work has shown that medical VLMs systematically fail to localize anatomically relevant regions even when general-domain models can \cite{liu2026medical}.

Multi-task training has been shown to improve both accuracy and representation quality in vision-language models. Methods such as 12-in-1 \cite{Lu_2020_CVPR} and Molmo \cite{Deitke_2025_CVPR} demonstrated that joint training across heterogeneous vision-language tasks benefits each individual task, and multi-modal pretraining in medical imaging has produced consistent segmentation gains \cite{rui2025multimodalvisionpretrainingmedical}. Grounded VQA has also been explored in the surgical domain: Surgical-VQLA \cite{bai2023surgicalvqlatransformergatedvisionlanguage} jointly answers questions and regresses the relevant region, while EndoChat \cite{wang2025endochat} scales this to large-scale grounded conversation in endoscopic surgery. Both confirm the value of grounding supervision in endoscopic settings, but target intraoperative surgical scenes with dedicated localisation annotations. Applied to diagnostic GI endoscopy, prior work \cite{safwan2025multitasklearningvisuallygrounded} submitted to the Medico 2025 challenge \cite{Medico2025} showed that multi-task training of a Florence-2 model improved both VQA accuracy and localization, with the segmentation objective acting as a useful regularizer. However, that work relied on ClipSeg \cite{luddecke2022clipseg} pseudo-masks requiring per-category prompt engineering with noisy results on visually diverse findings, used a multi-stage explanation pipeline that made it difficult to isolate visual grounding as a clean training signal, and evaluated a single backbone.

We replace the prompt-engineered pseudo-mask step with a simpler approach: fine-tuning a classification head on a GI-domain pretrained backbone (GastroNet-5M \cite{JONG2026174}) and using Grad-CAM \cite{Selvaraju_2019} to produce coarse region-of-interest masks. Because the classifier is trained on the same finding labels already present in the VQA dataset, its gradient maps are semantically grounded in GI findings without any per-category prompt design. For polyps, where high-quality expert masks already exist in Kvasir-SEG \cite{jha2020kvasir}, we continue to use them directly. We also replace the multi-stage explanation pipeline with Gemma-27B generating terminology-free visual descriptions of the relevant image region, which the training models then learn to reproduce when prompted to describe a finding they have answered. We evaluate across three small VLM backbones and additionally analyze the implicit cross-modal alignment in the model's internal representations, going beyond the explicit segmentation outputs.

To the best of our knowledge, this is the first work to evaluate multi-task grounded fine-tuning across multiple sub-2B VLMs for GI VQA and to analyze the resulting implicit cross-modal alignment as a measure of clinical trustworthiness. Concretely, we contribute:
\begin{enumerate}
    \item A multi-task fine-tuning recipe combining expert masks where available with Grad-CAM-based weak localization and a terminology-free visual description task, requiring no additional pixel-level annotation.
    \item An evaluation across three small VLMs (Florence-2, Qwen3.5, InternVL3.5) showing consistent VQA accuracy gains concentrated in appearance and spatial question types and increasing with question complexity.
    \item A grounding analysis showing that multi-task training improves implicit cross-modal alignment between answer tokens and relevant image regions, validated on in-distribution and out-of-distribution data with a paired significance test.
\end{enumerate}

\section{Methodology}

\begin{figure*}[t]
\centering
\includegraphics[width=\textwidth]{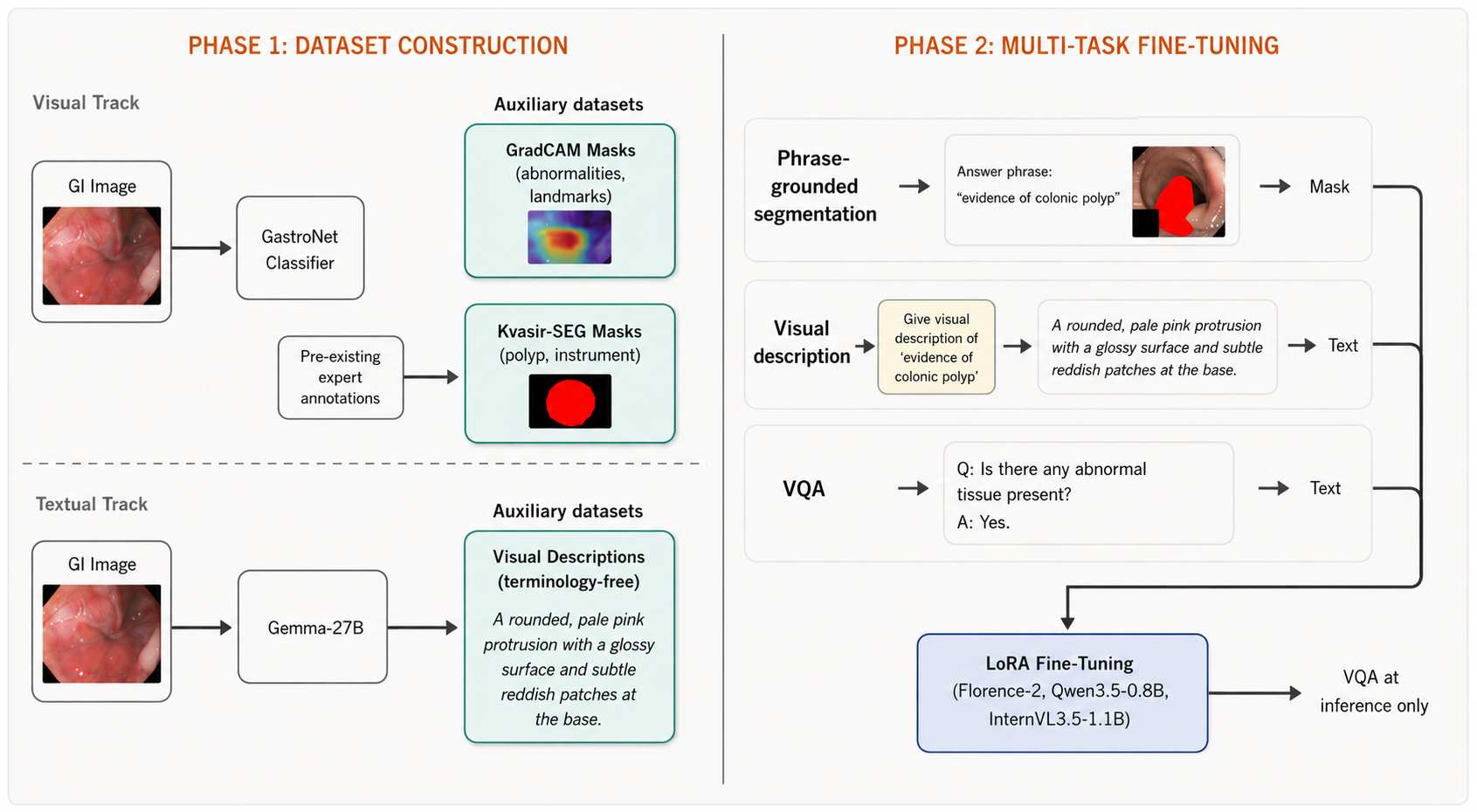}
\caption{Methodology overview. \textbf{Phase 1} constructs auxiliary datasets via two tracks: GradCAM masks from a GastroNet-5M classifier (abnormalities/landmarks), Kvasir-SEG expert masks (polyps), and Gemma-27B terminology-free visual descriptions. \textbf{Phase 2} fine-tunes each backbone jointly on three tasks via QLoRA; only VQA is used at inference.}
\label{fig:methodology}
\end{figure*}

Figure~\ref{fig:methodology} provides an overview of the full pipeline. \subsection{Region of Interest Mask Generation}
\label{sec:weakmask}

\paragraph{Polyps.} Expert-annotated segmentation masks are available from Kvasir-SEG \cite{jha2020kvasir}, and we reuse them directly. No weak supervision is needed for this category.

\paragraph{Abnormalities and landmarks.} No expert masks exist for these categories. We fine-tune a classification head (MLP) on top of a ResNet-50 initialized from GastroNet-5M pretrained weights \cite{JONG2026174}, keeping the feature extractor frozen. The classifier is trained on the Kvasir multi-class dataset \cite{pogorelov2017kvasir}, covering five GI finding classes (polyp, ulcerative colitis, esophagitis, cecum, z-line), for 30 epochs with Adam ($\text{lr}{=}5{\times}10^{-3}$, weight decay $10^{-4}$) and a step learning rate schedule. For each image we apply Grad-CAM \cite{Selvaraju_2019} at the final convolutional layer of ResNet-50 with respect to the predicted class $c$:
\begin{equation}
L^c = \mathrm{ReLU}\!\left(\sum_k \alpha^c_k A^k\right), \quad \alpha^c_k = \frac{1}{Z}\sum_{i,j}\frac{\partial y^c}{\partial A^k_{ij}}
\end{equation}
where $A^k$ is the $k$-th feature map of the final convolutional layer and $y^c$ is the classifier logit for class $c$. We retain only images where the classifier prediction probability exceeds 0.8 to filter unreliable activations, then apply Otsu thresholding to $L^c$. The mask is post-processed with morphological opening and closing (kernel size 5) and the largest connected component is kept; masks smaller than 100 pixels are discarded. GradCAM masks for polyp are not used since Kvasir-SEG expert annotations cover that class.

\subsection{Auxiliary Task Construction}

\paragraph{Phrase-grounded segmentation.} Each QA pair contains a finding-level answer (e.g.\ ``evidence of colonic polyp''). This answer phrase is used as the grounding target: the answer is linked to the corresponding mask (GradCAM-derived for abnormalities and landmarks, Kvasir-SEG for polyps), and the model is trained to localize the image region described by whichever answer phrase it produces. Multiple questions about the same finding map to the same mask. Florence-2 supports polygon/mask output via its native location tokens, so it is trained with full segmentation targets; Qwen3.5 and InternVL3.5 use bounding boxes, which their location token formats natively support. The final dataset contains 5,653 samples.

\paragraph{Visual description.} Gemma-27B \cite{gemmateam2025gemma3} generates a short visual description for each relevant image region, restricted to observable shape, texture, and color cues with medical terminology explicitly excluded. These descriptions serve as training targets: the VLMs are fine-tuned to produce the corresponding Gemma-generated description when given the task prompt \textit{``Give visual description of `\{answer\}' ''}, where \textit{answer} is the VQA answer for that image. This isolates visual-attribute learning as a clean signal separate from clinical language generation, and yields 3,299 training samples.

\subsection{Training}

Each backbone---Florence-2 (0.2B) \cite{xiao2023florence2}, Qwen3.5 (0.8B) \cite{qwen35blog}, and InternVL3.5 (1.1B) \cite{wang2025internvl35}---is fine-tuned with QLoRA \cite{dettmers2023qlora} for one epoch under two matched recipes: VQA-only (question-answer pairs only) and multi-task (MT, all three auxiliary tasks). Florence-2 uses LoRA rank~128, $\alpha{=}256$, lr~$5{\times}10^{-5}$, effective batch~8, trained with FP16 via the Transformers library. Qwen3.5 and InternVL3.5 use rank~8, $\alpha{=}32$, lr~$10^{-4}$, effective batch~16, trained with BFloat16 via ms-swift \cite{zhao2025swift}. All experiments run on a single NVIDIA RTX 3090 Ti.

\section{Experimental Setup}

The Kvasir-VQA-x1 dataset \cite{Gautam2025Jun} contains around 140k question-answer pairs. Due to resource constraints, we trained on a 50k subset and evaluated on a held-out 5k set. Both subsets were sampled to preserve the original distribution across question categories. All models were trained for one epoch. Each generated answer is scored by DeepSeek-V4 \cite{deepseekai2026deepseekv4} using a structured rubric that awards partial credit: multi-item answers score 1.0 if all ground-truth items are present, 0.5 for partial coverage, and 0.0 if entirely wrong; count questions score 1.0 for exact match and 0.5 if off by one; all other question classes are binary. The judge has no access to the image and evaluates strictly by comparing model output to the ground-truth answer text.

For the grounding analysis we use two additional datasets: 1000 Kvasir-VQA-x1 test images evaluated against Kvasir-SEG polyp masks (in-distribution), and all 612 CVC-ClinicDB \cite{BERNAL20123166} images evaluated against their expert polyp masks (out-of-distribution, never seen in training).

\section{Results}

\subsection{VQA Accuracy}

Multi-task training consistently improves overall accuracy across all three backbones: +2.74pp for Florence-2 (79.26 to 82.00), +1.53pp for InternVL3.5 (80.09 to 81.62), and +1.37pp for Qwen3.5 (80.35 to 81.72).

Kvasir-VQA-x1 categorizes each question into one of three complexity tiers: C1 answers one aspect of a finding, C2 two aspects simultaneously, and C3 three. Table~\ref{tab:complexity} reports mean accuracy per tier. For Florence-2, gains increase monotonically with complexity (C1: +1.31, C2: +2.78, C3: +3.36), consistent with multi-task training helping most when the model must address multiple visual attributes of a finding in a single response. InternVL3.5 shows the same directional trend at C1 and C3 (+1.32, +2.87). Qwen3.5 gains are smaller and do not follow this pattern clearly, which we discuss in Section~\ref{sec:conclusion}.

\begin{table}[t]
\centering
\caption{Mean accuracy (\%) by complexity tier. B.~=~Baseline (VQA-only), P.~=~Proposed (MT).}
\label{tab:complexity}
\setlength{\tabcolsep}{4pt}
\begin{tabular}{lcccccc}
\toprule
 & \multicolumn{2}{c}{C1} & \multicolumn{2}{c}{C2} & \multicolumn{2}{c}{C3} \\
\cmidrule(lr){2-3}\cmidrule(lr){4-5}\cmidrule(lr){6-7}
Model & VQA (B.) & MT (P.) & VQA (B.) & MT (P.) & VQA (B.) & MT (P.) \\
\midrule
Florence-2  & 86.0 & 87.3 & 77.4 & 80.2 & 75.1 & \textbf{78.4} \\
InternVL3.5 & 86.3 & 87.6 & 77.3 & 78.1 & 76.0 & \textbf{78.9} \\
Qwen3.5     & 87.0 & 86.6 & 79.1 & 80.0 & 77.3 & 77.4 \\
\bottomrule
\end{tabular}
\end{table}

We further group the 18 question categories into three buckets. \textbf{Appearance} questions ask about visual attributes (polyp size, polyp type, abnormality color, landmark color). \textbf{Spatial} questions ask where a finding is (abnormality location, landmark location, instrument location). \textbf{Structural} questions ask about category membership or discrete counts, which do not depend on fine-grained visual reasoning. Table~\ref{tab:stratified} reports the mean accuracy difference within each bucket.

\begin{table}[t]
\centering
\caption{Mean accuracy difference (MT$-$VQA-only, pp) by question type.}
\label{tab:stratified}
\begin{tabular}{lcccc}
\toprule
Model & Params & Appearance & Spatial & Structural \\
\midrule
Florence-2  & 0.2B & \textbf{+7.66} & +4.32 & +0.83 \\
InternVL3.5 & 1.1B & \textbf{+2.83} & +1.86 & +1.04 \\
Qwen3.5     & 0.8B & +0.95 & +1.70 & +1.53 \\
\bottomrule
\end{tabular}
\end{table}

For Florence-2 and InternVL3.5, the ordering appearance $>$ spatial $>$ structural holds, consistent with the auxiliary tasks teaching visual-attribute and localization reasoning. The largest individual gains are on polyp size (+12.1 for Florence-2) and polyp type (+8.4 Florence-2, +6.7 InternVL3.5), both of which require distinguishing fine-grained visual attributes. Table~\ref{tab:perclass} shows the full per-class breakdown.

\begin{table}[t]
\centering
\caption{Per-class VQA accuracy: Baseline$\to$Proposed (\boldmath$\Delta$, pp). Landmark color ($n=11$) excluded. Abn.=abnormality, Lmk.=landmark, Inst.=instrument.}
\label{tab:perclass}
\setlength{\tabcolsep}{2pt}
\scriptsize
\begin{tabular}{llccc}
\toprule
Bucket & Question class & Florence-2 & Qwen3.5 & InternVL3.5 \\
\midrule
\multirow{3}{*}{Appearance}
 & polyp size     & 65.9$\to$78.0 (\textbf{+12.1}) & 75.6$\to$78.7 (+3.1) & 74.3$\to$75.4 (+1.1) \\
 & polyp type     & 67.1$\to$75.5 (\textbf{+8.4})  & 69.4$\to$70.4 (+1.0) & 68.2$\to$74.9 (\textbf{+6.7}) \\
 & abn.\ color    & 60.8$\to$62.6 (+1.8)  & 64.3$\to$62.7 ($-$1.6) & 64.3$\to$64.7 (+0.4) \\
\midrule
\multirow{3}{*}{Spatial}
 & abn.\ location  & 29.1$\to$36.2 (\textbf{+7.1}) & 39.8$\to$40.7 (+0.9) & 38.4$\to$42.1 (+3.7) \\
 & inst.\ location & 82.7$\to$85.5 (+2.8) & 83.9$\to$85.5 (+1.6) & 83.9$\to$85.1 (+1.2) \\
 & lmk.\ location  & 69.1$\to$71.8 (+2.7) & 69.9$\to$72.7 (+2.8) & 68.8$\to$69.2 (+0.4) \\
\midrule
\multirow{6}{*}{Structural}
 & abn.\ presence  & 77.2$\to$80.0 (+2.8) & 71.6$\to$77.8 (\textbf{+6.2}) & 71.8$\to$76.8 (\textbf{+5.0}) \\
 & polyp removal   & 77.2$\to$81.0 (+3.8) & 80.2$\to$83.2 (+3.0) & 79.7$\to$81.4 (+1.7) \\
 & finding count   & 84.3$\to$84.6 (+0.3) & 81.0$\to$81.5 (+0.5) & 79.9$\to$81.8 (+1.9) \\
 & inst.\ count    & 96.7$\to$97.3 (+0.6) & 96.5$\to$96.2 ($-$0.3) & 95.7$\to$96.5 (+0.8) \\
 & text presence   & 84.6$\to$83.4 ($-$1.2) & 83.3$\to$85.5 (+2.2) & 84.7$\to$84.4 ($-$0.3) \\
 & procedure type  & 99.2$\to$99.4 (+0.2) & 97.5$\to$97.5 (0.0) & 98.3$\to$97.9 ($-$0.4) \\
\bottomrule
\end{tabular}
\end{table}

\subsection{Grounding Analysis}
\label{sec:grounding}

The analysis below probes the model's \emph{implicit} internal representations rather than its explicit location token outputs: specifically, whether multi-task training causes decoder answer-token embeddings to become more semantically aligned with the corresponding encoder visual patches. This tests whether grounding supervision propagates to improve cross-modal alignment in ordinary answer tokens, beyond dedicated output heads.

We compute per-token decoder-to-encoder-patch cosine similarity for each content token in the generated answer:
\begin{equation}
s(t_i, v_j) = \frac{\mathbf{e}_{t_i} \cdot \mathbf{e}_{v_j}}{\|\mathbf{e}_{t_i}\|\,\|\mathbf{e}_{v_j}\|}
\end{equation}
where $\mathbf{e}_{t_i}$ is the decoder embedding of the $i$-th answer token and $\mathbf{e}_{v_j}$ is the encoder patch embedding at position $j$. Pointing-game accuracy (PG) \cite{zhang2018top}, adapted to use the centroid of the top-10\% highest-similarity patches, reports the fraction of images where this centroid falls inside the ground-truth mask. Concentration ratio (CR) is:
\begin{equation}
\mathrm{CR}(t_i) = \frac{\frac{1}{|\mathcal{M}|}\sum_{j \in \mathcal{M}} s(t_i,v_j)}{\frac{1}{|\bar{\mathcal{M}}|}\sum_{j \notin \mathcal{M}} s(t_i,v_j)}
\end{equation}
where $\mathcal{M}$ is the set of patches inside the ground-truth mask; values near 1 indicate no systematic alignment. We compare VQA-only against full MT and two single-task ablations: \textbf{Vis} (segmentation only, no description) and \textbf{Des} (description only, no segmentation). Table~\ref{tab:grounding} reports all four conditions.

\begin{table}[t]
\centering
\caption{Token-to-patch grounding metrics, Florence-2. PG = pointing-game accuracy, CR = concentration ratio. Wilcoxon signed-rank (MT vs VQA-only, CVC-ClinicDB CR): $p = 1.47 \times 10^{-36}$.}
\label{tab:grounding}
\begin{tabular}{lcccc}
\toprule
 & \multicolumn{2}{c}{Kvasir (in-dist.)} & \multicolumn{2}{c}{CVC-ClinicDB (OOD)} \\
\cmidrule(lr){2-3} \cmidrule(lr){4-5}
 & PG & CR & PG & CR \\
\midrule
VQA (Baseline) & 0.333 & 1.404 & 0.142 & 1.285 \\
Vis only  & 0.734 & 1.664 & \textbf{0.727} & 1.540 \\
Des only  & 0.757 & \textbf{1.924} & 0.458 & 1.541 \\
MT (Proposed) & \textbf{0.805} & 1.920 & 0.645 & \textbf{1.576} \\
\bottomrule
\end{tabular}
\end{table}

All three auxiliary task conditions substantially improve over VQA-only on both metrics and both datasets. The ablation reveals a complementary pattern: Des alone almost matches full MT on in-distribution concentration ratio (1.924 vs 1.920), making the visual description task the primary driver of in-distribution cross-modal alignment. Vis alone, however, outperforms Des alone on out-of-distribution pointing-game accuracy (0.727 vs 0.458), indicating the segmentation supervision generalizes more robustly to unseen image sources. Full MT achieves the best or near-best result in every condition by combining both signals.

\subsection{Qualitative Analysis}

\begin{figure*}[t]
\centering
\includegraphics[width=\textwidth]{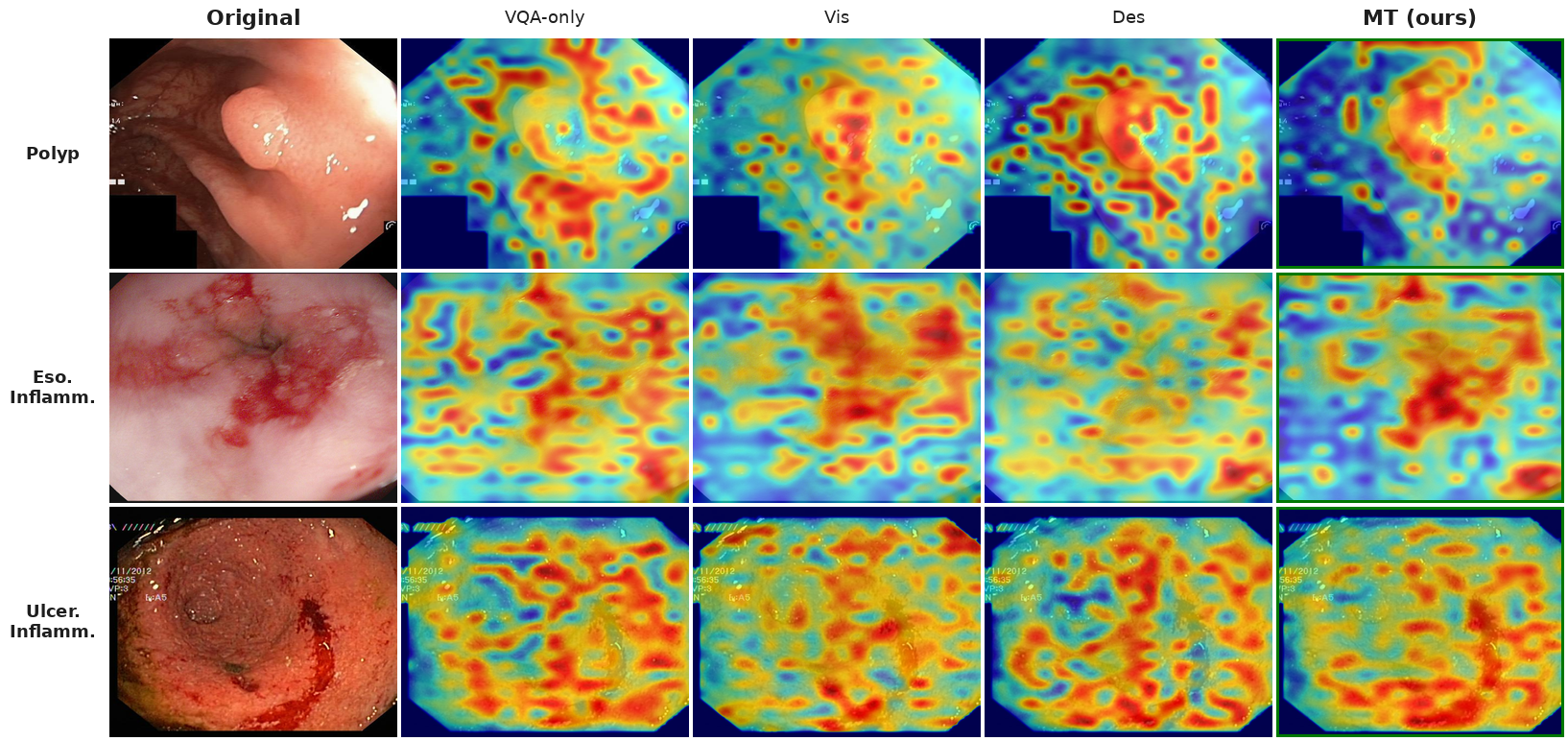}
\caption{Token-to-patch similarity maps (columns: original, VQA-only, Vis-only, Des-only, MT ours). Row 1: \texttt{polyp} token; rows 2--3: \texttt{inflammation} token. Des localizes the polyp tightly; Vis produces the most structured response for diffuse inflammation; MT is the least noisy overall. No ground-truth mask exists for inflammation categories.}
\label{fig:tokensim}
\end{figure*}

Figure~\ref{fig:tokensim} shows token-to-patch similarity maps across all four model variants for three cases: a polyp (discrete object) and two diffuse inflammation findings (esophageal and ulcerative). Across all three cases, MT consistently produces the least noisy activation maps, combining the tight localization of Des on discrete objects with the spatially structured response of Vis on diffuse findings.

For the polyp, all three auxiliary task conditions improve over VQA-only, and notably produce less noisy activation maps overall. Des alone produces the tightest localization, concentrating activation directly on the polyp body while suppressing surrounding tissue, consistent with the visual description task training the model to identify the specific object its answer refers to. MT produces a comparably tight map.

For both inflammation findings, which manifest as distributed texture and color changes without a hard boundary, VQA-only produces scattered, noisy activation with no spatial structure. Vis produces a noticeably more structured response, concentrating activation over the visually inflamed regions. Des alone, despite its strong performance on the polyp, reverts to noisy diffuse maps for these categories, similar to VQA-only. This mirrors the quantitative ablation in Table~\ref{tab:grounding}: Des drives in-distribution alignment for discrete objects but does not generalize to diffuse findings the way Vis does. No ground-truth mask exists for these inflammation categories; the assessment is purely by visual inspection of the implicit cross-modal alignment.

\section{Conclusion}
\label{sec:conclusion}

We proposed a simple multi-task fine-tuning recipe for GI VQA that reuses expert-annotated polyp masks and adds Grad-CAM weak supervision for finding categories without ground-truth masks, alongside a terminology-free visual description task. Applied to three small VLM backbones on a single GPU, the recipe produces consistent VQA accuracy gains concentrated in appearance and spatial question types, growing with question complexity. A grounding analysis confirms improvements extend to implicit cross-modal representations. The benefit is most pronounced for Florence-2 and InternVL3.5; Qwen3.5 shows smaller, flatter gains that may relate to architectural differences in spatial representation. Future work includes validating Grad-CAM mask quality with a spot-check, extending to additional finding categories, and running a controlled per-task ablation.


\subsubsection*{Disclosure of Interests.}
The authors have no competing interests to declare that are relevant to the content of this article.

\subsubsection*{Disclosure of AI Usage.}
Claude (Anthropic) assisted with writing and formatting; all scientific content, results, and conclusions are the authors' own.

\end{document}